\documentclass[runningheads]{llncs}
\usepackage[T1]{fontenc}
\usepackage{graphicx}
\usepackage{booktabs}
\usepackage{amsmath}
\usepackage{color}
\usepackage{float}
\usepackage[hyphens]{url}
\usepackage{hyperref}
\usepackage{tabularx}

\begin{document}
\title{Disease Burden over Skin Tone: Decomposing the Dermatology-AI Generalization Gap}
\titlerunning{Decomposing the Dermatology-AI Generalization Gap}
%
\authorrunning{Kunwor et al.}
\author{Nirajan Kunwor\textsuperscript{1} \and
Sanjaya Poudel\textsuperscript{2} \and
 Quoc-Huy Trinh\textsuperscript{3} \and
Jahidul Arafat\textsuperscript{4} \and
Sunil Kumar Gaire\textsuperscript{2*}}

\institute{\textsuperscript{1}Tribhuvan University, \textsuperscript{2}North Carolina A\&T State University, \textsuperscript{3}Aalto University, \textsuperscript{4}Auburn University
\\
{\small{*skgaire@ncat.edu}}
}
\maketitle
\begin{abstract}
Dermatology artificial intelligence (AI) models are trained predominantly on light-skinned, cancer-focused image collections, yet are increasingly proposed for deployment in resource-constrained settings (RCS) whose patients differ from the training population along two confounded axes at once: skin tone and disease distribution. When such models underperform, the field has largely attributed the failure to skin-tone underrepresentation. We evaluate a cancer-trained baseline (ResNet-50 fine-tuned on HAM10000 and ISIC~2019), two open dermatology foundation models (DermLIP, MONET), and a general-purpose vision model (DINOv3) as frozen feature extractors, probing them on a tone-stratified but disease-matched dataset (Diverse Dermatology Images, DDI) and a disease-shifted, tone-diverse dataset (Skin Condition Image Network, SCIN). DDI and SCIN are US-sourced datasets used here as controlled proxies for the two axes of shift; they are not RCS-collected cohorts, and our deployment-oriented conclusions are correspondingly inferential. In our setting, disease-distribution shift contributes more than skin tone: the cancer baseline degrades sharply from its in-domain performance to unfamiliar conditions (a drop from 0.62 to 0.21 balanced accuracy across clinical categories), a far larger effect than the within-disease skin-tone gap, which is comparatively small (0.10-0.18) and did not show a consistent direction. Using a label-free representation-quality analysis, we show this collapse is \emph{representational}, not merely an artifact of disjoint label spaces: cancer-specialized features barely cluster unfamiliar conditions (kNN neighbor-purity lift $+0.06$ over chance), whereas dermatology-pretrained features retain substantially more transferable structure ($+0.23$). Finally, representation quality closely tracks recoverable performance under cheap adaptation: starting from a dermatology foundation model, roughly ten labeled examples per clinical category recover most attainable performance. We release code and the evaluation protocol to support reproducible auditing. Code is available at: https://github.com/Nirajan995/dermatology-generalization-gap 

\end{abstract}
\section{Introduction}

Artificial intelligence (AI) for skin-disease classification has advanced rapidly, driven by large public archives such as ISIC and HAM10000. These archives are, however, narrow in two aspects: they are dominated by lighter skin tones and by neoplastic (cancer-related) conditions captured with dermoscopy. The settings where AI-assisted dermatology could have the greatest impact-resource-constrained primary-care and teledermatology clinics-differ from this training regime on both counts simultaneously. Patients present with more pigmented skin, \emph{and} with a disease burden dominated by inflammatory and infectious conditions (eczema, tinea, impetigo, scabies) that are largely absent from cancer-centric training data.

When dermatology AI generalizes poorly to these settings, the prevailing explanation emphasizes skin-tone underrepresentation, motivating substantial effort toward tone-diverse data collection~\cite{ref_ddi,ref_groh,ref_daneshjou_nm}. Yet skin tone and disease distribution are confounded in real deployments: a model that fails on a dark-skinned patient with eczema may fail because of skin tone, because eczema was never in its training set, or both. Distinguishing these causes is not merely academic; it determines where scarce data-collection and model-development resources should be directed. If the dominant driver is disease distribution, tone-diversification alone will not close the gap. Distribution shift is a known failure mode for medical AI~\cite{ref_wilds,ref_taori,ref_zech,ref_kelly}, and in underserved populations, it can translate directly into diagnostic bias~\cite{ref_seyyed}; dermatology is especially challenging because tone and disease prevalence shift simultaneously, obscuring the cause of failure.

Prior work has begun to probe this question. Rikhye et al.~\cite{ref_rikhye} found that errors in a proprietary teledermatology deployment tracked skin-condition categories more than demographics, but relied on non-public data and a relatively mild within-system shift. Open benchmarking efforts~\cite{ref_groger,ref_ddi,ref_groh} establish that dermatology foundation models transfer imperfectly to pigmented populations. What remains untested, in an open and reproducible setting, is a direct decomposition of the gap into its tone and distribution components at an \emph{extreme} shift and, more fundamentally, whether observed failures reflect a genuine representational deficit or merely the absence of output labels for unfamiliar conditions.

To resolve these questions, we conduct an open and fully reproducible study using only public datasets and free-tier compute. Our contributions are threefold:
\begin{enumerate}
\item \textbf{An open, reproducible decomposition} of the gap into tone and distribution effects, using DDI (matched diagnoses, stratified Fitzpatrick tone) for the tone effect and SCIN (tone-diverse, non-neoplastic) for the distribution effect. Distribution shift dominates across all evaluated models.
\item \textbf{A label-free representational analysis} showing the distribution-driven collapse is not an artifact of disjoint label spaces: cancer-specialized features fail to cluster unfamiliar conditions even before any classifier is trained, whereas dermatology-pretrained features retain substantially more transferable structure.
\item \textbf{A low-compute adaptation benchmark} showing recoverable performance is bounded by pre-existing representation quality ($r=0.90$), and that a dermatology foundation model adapted with $\sim$10 labeled examples per category recovers most attainable performance without retraining.
\end{enumerate}
Throughout we adopt patient-level splits and report bootstrap confidence intervals. DDI and SCIN are US-sourced proxies rather than RCS-collected cohorts; we return to this in Section~\ref{sec:discussion}.

\section{Previous Work}

\textbf{Diversity and fairness in dermatology AI:} Following the demonstration that deep networks can reach dermatologist-level skin-cancer classification~\cite{ref_esteva}, a recognized limitation of dermatology AI has been uneven performance across skin tones, with several analyses reporting lower accuracy on darker Fitzpatrick types~\cite{ref_daneshjou_nm}. The DDI dataset~\cite{ref_ddi} enables tone-stratified evaluation with biopsy-confirmed labels, and benchmarks such as Fitzpatrick17k~\cite{ref_groh} quantify these disparities, which pooled analyses place at a few AUROC points; analogous underdiagnosis biases appear in other medical-imaging domains~\cite{ref_seyyed}. Our work does not dispute that a tone effect exists; our own measurements on DDI recover a small one. We ask instead whether it is the \emph{dominant} driver of the deployment gap, or whether disease-distribution shift matters more. Our tone analysis builds directly on DDI. 

\textbf{Distribution shift versus demographics:} Rikhye et al.~\cite{ref_rikhye} provided evidence that condition distribution, rather than demographics, drove observed error disparities in a proprietary teledermatology system. Our work is a deliberate open replication-and-extension of this finding: we use publicly available models and data, and push to an extreme shift (Western cancer $\rightarrow$ non-neoplastic conditions) while additionally testing the \emph{representational} basis of the effect.

\textbf{Dermatology foundation models and open benchmarking:} Vision-language and self-supervised pretraining (e.g., CLIP~\cite{ref_clip}, DINOv2~\cite{ref_dinov2}) underpins a new generation of transferable feature extractors, including medical models~\cite{ref_medclip,ref_biomedclip} and, in dermatology, models trained on large clinical corpora~\cite{ref_dermlip,ref_monet,ref_panderm}. Open benchmarking efforts~\cite{ref_groger,ref_xu} have evaluated such models with frozen features on public datasets including pigmented-skin collections. We share the frozen-feature methodology but differ in objective: rather than ranking models, we use frozen features to \emph{decompose the causes} of the generalization gap and to characterize when cheap adaptation can and cannot help.

\section{Methodology}

\subsection{Datasets and Their Roles}
We use four public datasets, each assigned a specific experimental role (Table~\ref{tab:datasets}).
\emph{Source domain (in-domain reference):} HAM10000~\cite{ref_ham} (10{,}015 dermoscopic images) and ISIC~2019~\cite{ref_isic} (25{,}331 images) are pooled and mapped to a shared eight-class neoplastic taxonomy to train and evaluate the cancer baseline on familiar data.
\emph{Tone-isolation.} DDI (656 images; Fitzpatrick groups I-II, III-IV, V-VI) holds diagnosis broadly fixed while varying skin tone, isolating the tone effect; we use its binary malignant/benign label.
\emph{Distribution-isolation.} SCIN~\cite{ref_scin} (6{,}517 images after cleaning; 3{,}061 patients) is tone-diverse but dominated by non-neoplastic conditions absent from cancer training data, isolating the distribution effect.

We note that DDI is tone-\emph{stratified} (Fitzpatrick groups in comparable numbers by construction) while SCIN is tone-\emph{diverse} but not tone-balanced. Our design therefore observes the two axes marginally rather than jointly; the tone effect \emph{under} shifted disease cannot be isolated in these datasets without confounding it with the change in label space, since DDI is scored on a binary task and SCIN on multi-class categories. Characterizing that interaction requires a cohort with tone and condition jointly annotated under one label space, which we leave to future work.

\begin{figure}[t]
    \centering
    \includegraphics[width=0.45\linewidth]{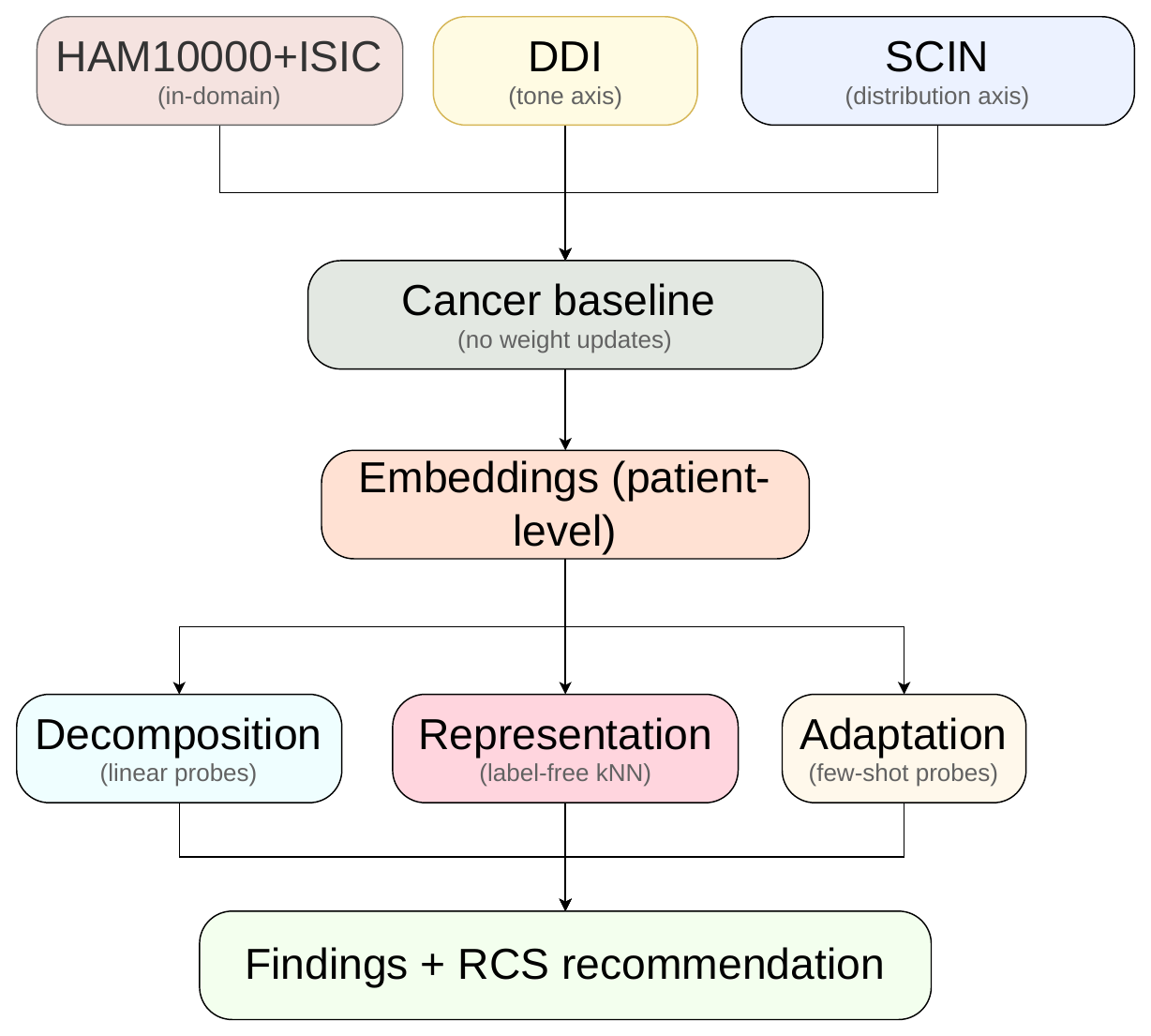}
    \caption{System architecture: four frozen models encode four public datasets into patient-level embeddings, evaluated by three complementary analyses.}
    \label{fig:arch}
\end{figure}

\begin{table}[t]
\centering
\caption{Datasets and their experimental roles. All evaluation uses patient-level splits.}
\label{tab:datasets}
\small
\setlength{\tabcolsep}{4pt}
\begin{tabularx}{\linewidth}{@{}l r l X@{}}
\toprule
\textbf{Dataset} & \textbf{Images} & \textbf{Domain} & \textbf{Role} \\
\midrule
HAM10000 + ISIC 2019 & 35{,}346 & Western, neoplastic & Source / in-domain reference \\
DDI & 656 & Tone-stratified & Isolate tone effect \\
SCIN & 6{,}517 & Tone-diverse, non-neoplastic & Isolate distribution effect \\
\bottomrule
\end{tabularx}
\end{table}

\subsection{Models}
We evaluate four models as frozen feature extractors: (i) a \textbf{cancer baseline}, ResNet-50 (ImageNet-initialized) fine-tuned on the pooled HAM10000+ISIC eight-class task, with the classification head removed to yield 2048-d features; (ii) \textbf{DermLIP}~\cite{ref_dermlip} (512-d) and (iii) \textbf{MONET}~\cite{ref_monet} (1024-d), open dermatology vision-language foundation models; and (iv) \textbf{DINOv3}~\cite{ref_dinov3} (\path{facebook/dinov3-vitb16-pretrain-lvd1689m}, pooler output, 768-d), a general-purpose self-supervised vision model pretrained on natural images, included both as a no-dermatology-knowledge control and as a representative of the general-purpose vision foundation models now routinely used as feature extractors in medical imaging. For reference, the remaining encoders are \path{hf-hub:redlessone/DermLIP_ViT-B-16} (image encoder, 512-d) and \texttt{chanwkim/monet} (vision-tower pooler output, 1024-d). No model weights are updated during evaluation.

\subsection{Evaluation Protocol}
For each (model, dataset) pair we extract frozen embeddings and evaluate with three analyses.
\emph{(1) Linear probing.} We fit an L2-regularized logistic-regression probe on standardized frozen features, using patient-level \texttt{GroupShuffleSplit} (70/30) to prevent leakage, and report balanced accuracy. For SCIN, we map its long-tailed conditions into seven clinically grounded categories (inflammatory, infectious, vascular/purpuric, neoplastic, pigmentary, traumatic/other, and other) following standard dermatological grouping; the category probe uses all seven, while the secondary label-free category analysis retains the five categories with at least 50 images. We report both category-level and fine-grained results. The cancer baseline is trained on HAM10000+ISIC with class-weighted cross-entropy, AdamW, and cosine scheduling, reaching 0.673 validation balanced accuracy.
\emph{(2) Label-free representation quality.} Independent of any classifier, we compute $k$-nearest-neighbor neighbor purity ($k=10$, cosine metric): the fraction of each image's neighbors sharing its label. We report purity minus the dataset's random-neighbor chance floor ($\sum_i p_i^2$), termed \emph{lift}, so imbalance cannot inflate the metric. This asks whether frozen features geometrically separate conditions \emph{before} any labels are used for classification.
\emph{(3) Low-compute adaptation.} On SCIN categories, we compare a full probe, a few-shot probe (10 examples/class, averaged over 5 draws), and raw-vs-standardized features, all without retraining.
Confidence intervals are 1000-sample bootstraps for probe accuracies and 500-sample bootstraps for purity.

\emph{Comparability across models.} Every stage downstream of feature extraction is identical across the four encoders: the same patient-level split (seed 42), the same standardization fitted on training data only, the same probe with identical solver, regularization and class weighting and no per-model tuning, the same $k$ and metric for purity, and the same bootstrap procedure. Feature extraction uses each encoder's native preprocessing (ImageNet normalization for the cancer baseline and DINOv3, CLIP for MONET, DermLIP's own transform) at a common $224\times224$ resolution; imposing one normalization would evaluate three models off their pretraining distribution. Embedding dimensionality is not equalized, and we return to its effect in Section~\ref{sec:discussion}.

\emph{Compute resources and timing:} All experiments ran on free-tier cloud compute (a single Kaggle NVIDIA T4 GPU session plus CPU-only sessions), deliberately matching resource-limited constraints. Baseline fine-tuning and frozen feature extraction for all four models across all datasets each completed in one GPU session; all downstream analyses are CPU-only and complete in minutes. At inference each image requires one forward pass through a frozen encoder plus a lightweight linear probe, so no specialized hardware is needed to reproduce our results.

\section{Results}

\subsection{Distribution Shift Dominates the Generalization Gap}
Absolute probe accuracies are modest by design: features are frozen, so they reflect linear separability of fixed representations rather than the ceiling of a trained model (our cancer baseline reaches 0.673 fine-tuned end-to-end). Our argument therefore rests on relative drops across domains and on the label-free analysis of Section~\ref{sec:repr}.

Table~\ref{tab:main} contrasts in-domain performance, the distribution effect (SCIN), and the tone effect (DDI). The cancer baseline is strong in-domain (0.62) but degrades sharply on unfamiliar conditions (0.21 across categories; 0.06 at fine granularity). Dermatology foundation models degrade far less (DermLIP 0.36), and even the no-dermatology control (DINOv3) transfers better than the cancer baseline. By contrast, the DDI tone gap is small (0.10-0.18) and \emph{inconsistent in direction}; the cancer baseline's worst group is the lightest, DINOv3's the darkest, indicating no systematic monotonic tone bias.

\begin{table}[t]
\centering
\caption{The generalization gap decomposed. In-domain and SCIN columns report
full-probe balanced accuracy; the DDI tone gap is the max--min across Fitzpatrick
groups with 95\% bootstrap confidence intervals (CIs). Distribution shift produces a large drop; the tone
gap is smaller and inconsistent in direction.}\label{tab:main}
\begin{tabular*}{\linewidth}{@{\extracolsep{\fill}}lccc@{}}
\toprule
\textbf{Model} & \textbf{In-domain} & \textbf{SCIN (distribution)} & \textbf{DDI tone gap} \\
\midrule
Cancer baseline & 0.62 & 0.21 & 0.14 (0.03-0.28) \\
DermLIP         & 0.52 & 0.36 & 0.14 (0.03-0.26) \\
MONET           & 0.52 & 0.34 & 0.10 (0.02-0.20) \\
DINOv3          & 0.47 & 0.31 & 0.18 (0.05-0.31) \\
\bottomrule
\end{tabular*}
\end{table}

\subsection{Evidence for a Representational, Not Definitional, Gap}\label{sec:repr}
\begin{figure}[htbp]
\centering
\includegraphics[width=0.7\linewidth]{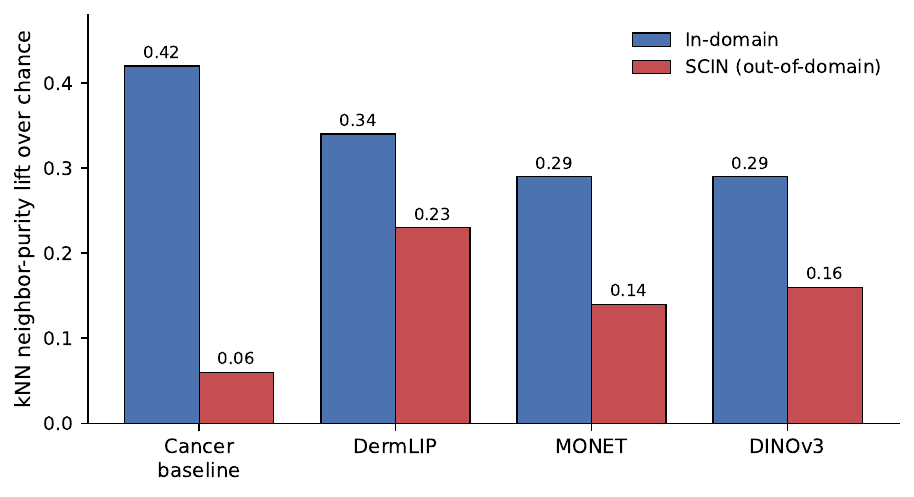}
\caption{Label-free representation quality (kNN purity lift over chance), in-domain versus SCIN. The cancer baseline is strongest in-domain but collapses to near-chance on unfamiliar conditions.}\label{fig:collapse}
\end{figure}

A natural objection is that low accuracy on unfamiliar diseases is trivially expected, since the cancer baseline was never trained on those labels. Our label-free analysis refutes this (Fig.~\ref{fig:collapse}). On its home domain, the cancer baseline has the strongest representational structure of any model (purity lift $+0.42$), yet on SCIN its features barely separate conditions ($+0.06$) close to chance, demonstrating a genuine representational deficit rather than a missing output head. Dermatology foundation models retain three to four times more transferable structure under the same unfamiliar conditions (DermLIP $+0.23$), and the general-vision control also exceeds the cancer baseline. Thus, specialization purchases in-domain strength at the direct cost of transferable structure.

DINOv3's behavior deserves comment: with no dermatology pretraining it nonetheless transfers to SCIN better than the cancer baseline (0.31 vs.\ 0.21 balanced accuracy; $+0.16$ vs.\ $+0.06$ purity lift) while being the weakest model in-domain (0.47). If the gap were caused by absent dermatology knowledge, a natural-image model should transfer worst, not second-best. The ordering instead tracks the \emph{breadth} of pretraining relative to the target: narrow supervised fine-tuning discards exactly the variation that is uninformative for melanoma-versus-nevus and essential for eczema-versus-tinea. Specialization is better read as a lossy projection than as accumulated knowledge. A systematic comparison across general-vision foundation models would sharpen this and is left to future work.

\subsection{Cheap Adaptation Is Bounded by Latent Structure}
Table~\ref{tab:adapt} reports low-compute adaptation on SCIN categories. Across models, recoverable performance closely tracks the label-free representation quality of Fig.~\ref{fig:collapse} ($r=0.90$ between SCIN purity lift and full-probe balanced accuracy across the four models; suggestive given $n=4$, not a significance test). The cancer baseline, lacking latent structure, recovers least (0.21) even with adaptation, whereas dermatology models recover substantially more, and for these models a few-shot probe with only ten labeled examples per category nearly matches or exceeds the full probe (DermLIP: 0.41 vs.\ 0.36; MONET: 0.39 vs.\ 0.34). Feature standardization alone provided no consistent benefit. The pattern suggests a promising direction for RCS deployment: the choice of starting representation appears to matter more than the adaptation method, and a dermatology foundation model may be adaptable to local conditions with relatively little labeled data. We frame this as a potential rather than a validated recipe; confirming it at scale is left to future work.

\begin{table}[h]
\caption{Low-compute adaptation on SCIN categories (balanced accuracy). Few-shot uses 10 examples/class. Recovery tracks latent structure (Fig.~\ref{fig:collapse}); the cancer baseline cannot be cheaply rescued.}\label{tab:adapt}
\centering
\begin{tabular*}{\linewidth}{@{\extracolsep{\fill}}lccc@{}}
\toprule
\textbf{Model} & \textbf{Full probe} & \textbf{Few-shot ($k=10$)} & \textbf{SCIN purity lift} \\
\midrule
Cancer baseline & 0.21 & 0.19 & +0.06 \\
DermLIP         & 0.36 & 0.41 & +0.23 \\
MONET           & 0.34 & 0.39 & +0.14 \\
DINOv3          & 0.31 & 0.32 & +0.16 \\
\bottomrule
\end{tabular*}
\end{table}

\section{Discussion}\label{sec:discussion}

Our results give a clear ordering of causes for the generalization gap in the settings studied: disease-distribution shift is a large, representationally grounded driver, whereas the matched-disease tone effect is small and inconsistent in direction. The tight coupling between latent structure and cheap recoverability suggests a practical RCS principle: invest in the starting representation rather than in expensive local retraining. 

\textbf{Limitations.} First and most importantly, DDI and SCIN are US-sourced proxies used to isolate each factor, not RCS-collected data; our RCS conclusions are therefore inferential, and validation on genuinely RCS-collected cohorts is important future work. Second, the DDI tone analysis is statistically underpowered (55-77 images per Fitzpatrick group in the test split); we therefore avoid strong directional claims. Third, the adaptation correlation is computed over four models and is descriptive rather than inferential. Fourth, we evaluate frozen features only; end-to-end fine-tuning may alter the picture, and parameter-efficient fine-tuning (e.g., LoRA) is now feasible even on free-tier compute and would be the natural next test of whether latent structure bounds recovery as tightly under partial weight updates as under linear probing. Fifth, the cancer baseline differs from the foundation models in architecture, pretraining objective and pretraining scale simultaneously, so specialization cannot be fully isolated from these factors. Finally, several factors bound interpretation: our accuracy comparison is not chance-normalized across tasks of differing class count (our chance-corrected label-free analysis is more directly comparable); kNN purity is not controlled for embedding dimensionality, and the four encoders differ in dimensionality by a factor of four; the mapping of SCIN's long tail into broad clinical categories simplifies genuine diagnostic complexity; Fitzpatrick type indexes UV response rather than pigmentation; and stronger dermatology foundation models (e.g., PanDerm~\cite{ref_panderm}) remain to be evaluated within this framework.

\textbf{Conclusion.} Using only public data and free-tier compute, we decomposed the dermatology-AI generalization gap and found disease-distribution shift to dominate skin tone, traced the effect to a representational deficit in cancer-specialized models, and found that cheap adaptation is most effective when the starting representation already encodes the target conditions.

\section{Impact in Resource-Constrained Settings}
Our findings suggest tentative guidance for teams building dermatology AI under resource constraints. First, \emph{model selection may matter more than local retraining}: since recoverable performance is bounded by representation quality, teams may benefit from starting with a dermatology-pretrained foundation model rather than a cancer-specialized classifier. Second, \emph{modest labeling may suffice}: a few-shot probe with roughly ten labeled examples per local category recovered most attainable performance in our experiments, suggesting a clinic could adapt a model with a small, locally curated label set. Third, the adaptation step itself is lightweight and runs on CPU, so the main hardware demand falls on the one-time feature extraction rather than on repeated local training. Finally, while tone-diverse data remain important for fairness auditing, closing the deployment gap under a different disease burden appears to require prioritizing \emph{disease-distribution coverage}.



\begin{thebibliography}{20}
\bibitem{ref_rikhye}
Rikhye, R.V., Loh, A., Hong, G.E., et al.: Closing the AI generalisation gap by adjusting for dermatology condition distribution differences across clinical settings. eBioMedicine \textbf{116}, 105766 (2025)

\bibitem{ref_groger}
Gröger, F., et al.: Towards scalable foundation models for digital dermatology. arXiv:2411.05514 (2024)

\bibitem{ref_ddi}
Daneshjou, R., Vodrahalli, K., Novoa, R.A., et al.: Disparities in dermatology AI performance on a diverse, curated clinical image set. Science Advances \textbf{8}(31), eabq6147 (2022)

\bibitem{ref_dermlip}
Yan, S., et al.: Derm1M: A million-scale vision-language dataset aligned with clinical ontology knowledge for dermatology. In: IEEE/CVF International Conference on Computer Vision (ICCV), pp. 12681--12690 (2025)

\bibitem{ref_monet}
Kim, C., Gadgil, S.U., DeGrave, A.J., Omiye, J.A., Cai, Z.R., Daneshjou, R., Lee, S.-I.: Transparent medical image AI via an image--text foundation model grounded in medical literature. Nature Medicine \textbf{30}(4), 1154--1165 (2024)

\bibitem{ref_dermacon}
Madarkar, S.S., et al.: DermaCon-IN: A multi-concept annotated dermatological image dataset of Indian skin disorders for clinical AI research. In: Advances in Neural Information Processing Systems (NeurIPS), Datasets and Benchmarks Track (2025)

\bibitem{ref_ham}
Tschandl, P., Rosendahl, C., Kittler, H.: The HAM10000 dataset, a large collection of multi-source dermatoscopic images of common pigmented skin lesions. Scientific Data \textbf{5}, 180161 (2018)

\bibitem{ref_scin}
Ward, A., et al.: Crowdsourcing dermatology images with Google Search ads: creating a real-world skin condition dataset (SCIN). arXiv:2402.18545 (2024)

\bibitem{ref_groh}
Groh, M., Harris, C., Soenksen, L., Lau, F., Han, R., Kim, A., Koochek, A., Badri, O.: Evaluating deep neural networks trained on clinical images in dermatology with the Fitzpatrick 17k dataset. In: IEEE/CVF Conference on Computer Vision and Pattern Recognition (CVPR) Workshops, pp. 1820--1828 (2021)

\bibitem{ref_isic}
Codella, N., Rotemberg, V., Tschandl, P., et al.: Skin lesion analysis toward melanoma detection 2018: A challenge hosted by the International Skin Imaging Collaboration (ISIC). arXiv:1902.03368 (2019)

\bibitem{ref_esteva}
Esteva, A., Kuprel, B., Novoa, R.A., Ko, J., Swetter, S.M., Blau, H.M., Thrun, S.: Dermatologist-level classification of skin cancer with deep neural networks. Nature \textbf{542}(7639), 115--118 (2017)

\bibitem{ref_daneshjou_nm}
Groh, M., Badri, O., Daneshjou, R., et al.: Deep learning-aided decision support for diagnosis of skin disease across skin tones. Nature Medicine \textbf{30}(2), 573--583 (2024)

\bibitem{ref_dinov3}
Siméoni, O., et al.: DINOv3. arXiv:2508.10104 (2025)

\bibitem{ref_xu}
Xu, S., Gui, H., Rotemberg, V., Wang, T., Chen, Y.T., Daneshjou, R.: A framework for evaluating the efficacy of foundation embedding models in healthcare. medRxiv 2024.04.17.24305983 (2024)

\bibitem{ref_wilds}
Koh, P.W., Sagawa, S., Marklund, H., et al.: WILDS: A benchmark of in-the-wild distribution shifts. In: International Conference on Machine Learning (ICML), pp. 5637--5664 (2021)

\bibitem{ref_taori}
Taori, R., Dave, A., Shankar, V., Carlini, N., Recht, B., Schmidt, L.: Measuring robustness to natural distribution shifts in image classification. In: Advances in Neural Information Processing Systems (NeurIPS) \textbf{33}, 18583--18599 (2020)

\bibitem{ref_dinov2}
Oquab, M., Darcet, T., Moutakanni, T., et al.: DINOv2: Learning robust visual features without supervision. Transactions on Machine Learning Research (TMLR) (2024)

\bibitem{ref_clip}
Radford, A., Kim, J.W., Hallacy, C., et al.: Learning transferable visual models from natural language supervision. In: International Conference on Machine Learning (ICML), pp. 8748--8763 (2021)

\bibitem{ref_biomedclip}
Zhang, S., Xu, Y., Usuyama, N., et al.: A multimodal biomedical foundation model trained from fifteen million image--text pairs. NEJM AI \textbf{2}(1), AIoa2400640 (2025)

\bibitem{ref_medclip}
Wang, Z., Wu, Z., Agarwal, D., Sun, J.: MedCLIP: Contrastive learning from unpaired medical images and text. In: Proceedings of the Conference on Empirical Methods in Natural Language Processing (EMNLP), pp. 3876--3887 (2022)

\bibitem{ref_zech}
Zech, J.R., Badgeley, M.A., Liu, M., Costa, A.B., Titano, J.J., Oermann, E.K.: Variable generalization performance of a deep learning model to detect pneumonia in chest radiographs: A cross-sectional study. PLoS Medicine \textbf{15}(11), e1002683 (2018)

\bibitem{ref_kelly}
Kelly, C.J., Karthikesalingam, A., Suleyman, M., Corrado, G., King, D.: Key challenges for delivering clinical impact with artificial intelligence. BMC Medicine \textbf{17}(1), 195 (2019)

\bibitem{ref_seyyed}
Seyyed-Kalantari, L., Zhang, H., McDermott, M.B.A., Chen, I.Y., Ghassemi, M.: Underdiagnosis bias of artificial intelligence algorithms applied to chest radiographs in under-served patient populations. Nature Medicine \textbf{27}(12), 2176--2182 (2021)

\bibitem{ref_panderm}
Yan, S., Yu, Z., Primiero, C., et al.: A multimodal vision foundation model for clinical dermatology. Nature Medicine \textbf{31}(8), 2691--2702 (2025)
\end{thebibliography}
\end{document}